\documentclass[11pt]{article}

\usepackage[]{acl}

\usepackage{times}
\usepackage{latexsym}
\usepackage{tikz}
\usepackage{tikz}
\usetikzlibrary{positioning,arrows.meta,calc}
\usepackage[T1]{fontenc}

\usepackage[utf8]{inputenc}

\usepackage{microtype}

\usepackage{inconsolata}

\usepackage{graphicx}

\title{
Towards End-to-End Multilingual Metaphor Processing: Integrating Detection, Translation, and Evaluation
}

\author{ Jiahui Liang$^1$, Lifeng Han$^{2,3}$ \\
  $^1$\texttt{Centre for Linguistics, Humanities,
  Leiden University, NL}
\\
  $^2$\texttt{LIACS, Leiden University, NL} \\
  $^3$\texttt{BDS, Leiden University Medical Centre, NL} \\
j.h.l.jiahui@hum.leidenuniv.nl | l.han@lumc.nl
 \\
 \\
}

\begin{document}
\maketitle
\begin{abstract}

Metaphorical language remains a major challenge for multilingual natural language processing because successful interpretation and translation require reasoning beyond literal lexical meaning. Existing research has largely investigated metaphor detection, machine translation, and translation evaluation as separate tasks, while little work has explored how these components can be integrated into a unified computational framework. This PhD proposal aims to develop an end-to-end framework for multilingual metaphor processing consisting of three complementary research directions: (1) robust metaphor detection across languages, (2) metaphor-oriented translation evaluation for both human assessment and automatic quality estimation, and (3) joint modelling that connects metaphor detection with translation evaluation. The proposed research will combine linguistic theory with recent advances in large language models to develop new datasets, annotation methodologies, evaluation benchmarks, and automatic evaluation approaches for metaphor-aware machine translation. The expected outcome is a unified framework that improves both the development and evaluation of multilingual NLP systems when processing figurative language.

\end{abstract}



\section{Introduction}
Metaphors are widespread in our daily usage and play a central role in human cognition and communication. Rather than being merely a rhetorical device used for stylistic effect, metaphor is regarded as a fundamental mechanism through which people conceptualise and understand the world \cite{johnson1980metaphors, kovecses2010metaphor}. Specifically, the process is realized by conceptualising the abstract target domain through mappings from the concrete source domain \cite{johnson1980metaphors}. For example, in the typical conceptual metaphor TIME IS MONEY, time is understood as a valuable resource.

These conceptual mappings are grounded in embodied physical and cultural experiences \cite{johnson1980metaphors, Gibbs2008}, while their interpretation often depends on contextual information \cite{li2025mindmachine} and shared cultural knowledge \cite{johnson1980metaphors}. Because embodied experiences and conceptualizations may differ across individuals and cultures \cite{kovecses2010metaphor}, metaphorical meaning is often non-compositional and cannot be derived from the literal meanings of individual words alone. Instead, it requires the integration of linguistic, conceptual, contextual, and cultural information. 
These challenges have motivated increasing research on computational metaphor processing, including metaphor detection, interpretation, generation and translation
\cite{rai2020survey, ge2023survey}. 
Among these tasks, metaphor \textbf{detection} and metaphor \textbf{translation} have developed largely as \textit{independent} research directions \cite{liang2025using,liang2026metahope}. 
Metaphor detection aims to distinguish metaphorical from literal language and serves as a prerequisite for many downstream applications \cite{shutova2010models}, whereas metaphor translation focuses on conveying metaphorical meaning across languages while accounting for linguistic and cultural differences \cite{karakanta2025metaphors}.
Despite substantial advances in neural language models, both tasks remain challenging. Successful metaphor detection requires models to recognize non-literal meaning beyond surface lexical patterns, while metaphor translation must further address variations in linguistic form, communicative function, and cultural embeddedness across languages \cite{karakanta2025metaphors}. 
However, relatively \textit{little work has connected these two areas}. Existing translation evaluation methods generally assume that metaphor instances have already been identified manually, while advances in metaphor detection have rarely been incorporated into translation evaluation or quality estimation.

Recent evidence also suggests that large language models may rely more heavily on superficial lexical cues than on genuine metaphor understanding, indicating that figurative language remains difficult even for state-of-the-art systems \cite{sanchez2025metaphor}. 
Furthermore, progress in metaphor translation is constrained by the limited availability of high-quality parallel metaphor corpora and systematic evaluation resources \cite{wang2024mmte,han2026towards}.

Recent work has also highlighted shortcomings in current evaluation practice. General-purpose MT metrics, such as BLEU \cite{banerjee-lavie-2005-meteor} or COMET \cite{rei-etal-2020-comet}, are not designed to capture metaphor-specific phenomena, and human evaluation protocols rarely distinguish different types of metaphor translation errors. Emerging resources, including MMTE \citep{wang2024mmte}, have begun addressing this gap, while our MetaHOPE framework extends the human-oriented HOPE evaluation framework with metaphor-specific error categories and annotation guidelines \cite{liang2026metahope}. Nevertheless, reliable metaphor-oriented evaluation and its integration with automatic metaphor processing remain open research problems.

This PhD proposes an end-to-end computational framework for multilingual metaphor processing that unifies three complementary research directions: (1) metaphor detection, (2) metaphor-oriented translation evaluation integrating MetaHOPE, and (3) joint modelling of metaphor detection and translation evaluation. Rather than treating these components as isolated tasks, the proposed research investigates how they can mutually reinforce one another to improve both the evaluation and the development of MT and LLM systems. The long-term objective is to establish reliable methodologies, resources and benchmarks for metaphor-aware multilingual NLP.

\section{Background and Related Work}

\subsection{Cognitive Metaphor}

Metaphor has traditionally been regarded as a rhetorical device used to enrich literary expression. However, this view was fundamentally challenged by Conceptual Metaphor Theory (CMT), which argues that metaphor is not merely a property of language but a fundamental mechanism of human thought and cognition \citep{johnson1980metaphors}. According to CMT, abstract concepts are systematically understood through more concrete source domains, giving rise to conceptual mappings such as \textit{ARGUMENT IS WAR} and \textit{TIME IS MONEY}. Linguistic metaphorical expressions are therefore surface manifestations of deeper conceptual structures rather than isolated stylistic devices.

Subsequent research has further demonstrated that metaphor is a multidimensional phenomenon involving cognition, language, communication, and culture. While conceptual mappings provide the cognitive basis for metaphorical expressions, their linguistic realisations are influenced by discourse context and communicative intention \cite{semino2008metaphor}. Deliberate Metaphor Theory (DMT), for example, distinguishes between metaphors that are intentionally employed to direct readers' attention towards the source domain and those that have become conventionalised in everyday language \citep{steen2017deliberate}. This distinction is particularly relevant for computational processing because not all metaphorical expressions require the same level of semantic interpretation.

Metaphor is especially prevalent in news discourse, political communication and other public-facing genres, where it serves multiple communicative functions beyond stylistic ornamentation. Previous studies have shown that metaphor can simplify complex events, frame public opinion, evoke emotional responses, and communicate ideological positions \citep{steen2010method}. Consequently, successful metaphor processing requires models to capture not only lexical meaning but also discourse-level and pragmatic information.

Cross-linguistic variation further increases the complexity of metaphor processing. Although some conceptual metaphors appear to be widely shared across cultures, their linguistic realisations often differ considerably because of language-specific conventions, cultural experiences, and socio-historical backgrounds \citep{kovecses2010metaphor}. As a result, multilingual NLP systems must account for both universal conceptual mappings and culture-specific metaphorical expressions \cite{han2025dutch}. 
These challenges motivate the need for computational approaches capable of recognising, interpreting, translating and evaluating metaphorical language across languages.

\subsection{Computational Metaphor Detection}

Automatic metaphor detection has become one of the most active research topics in figurative language processing \cite{shutova2015design,rai2020survey,han2025dutch}. Early computational approaches mainly relied on hand-coded knowledge, lexical resources, and selectional preferences to distinguish metaphorical from literal language \cite{shutova2015design}. However, these rule-based methods were often difficult to scale and generalise across domains and languages, motivating the development of data-driven approaches.



With the development of corpus-based metaphor research, manual metaphor identification procedures also became increasingly standardised. The Metaphor Identification Procedure (MIP) \citep{group2007mip} and its extended version, MIPVU \citep{steen2010method}, provide systematic dictionary-based guidelines for determining whether a lexical unit is used metaphorically by comparing its contextual meaning with a more basic meaning . Based on this procedure, the VU Amsterdam Metaphor Corpus (VUAMC) \cite{steen2010vu} was constructed as one of the largest manually annotated metaphor corpora and has since become a widely adopted benchmark for metaphor detection research. The availability of MIPVU-annotated corpora has enabled the development and evaluation of supervised metaphor detection models under consistent annotation standards.

Advances in deep learning have substantially improved metaphor detection performance. Contextual language models, including BERT and subsequent Transformer architectures, enable richer semantic representations by modelling contextual information surrounding metaphorical expressions. These models consistently outperform earlier statistical approaches on standard benchmark datasets and demonstrate improved robustness across multiple domains. This transition was reflected in the VUA Metaphor Detection Shared Tasks. While the top-performing systems in 2018 were mainly based on recurrent neural networks and feature engineering, the 2020 shared task was dominated by BERT-based models, with the best F1 score improving from 0.651 to 0.769 \cite{leong2018report, leong2020report}.

More recently, researchers have begun to explore large language models (LLMs) for automatic metaphor detection. Instead of relying exclusively on task-specific supervised training, LLMs can identify metaphorical expressions through zero-shot and few-shot prompting. 
Recent studies have investigated the use of Llama, Qwen, Mistral, DeepSeek, Gemma, 
GPT-4, and other LLMs for metaphor identification, demonstrating their potential for detecting conventional metaphorical expressions while also revealing limitations in achieving robust metaphor understanding \citep{liang2025using,reimann2025using,han2025dutch}. Building on these advances, recent work has explored task-specific prompt design, MIPVU-inspired reasoning, chain-of-thought prompting, and retrieval-augmented generation (RAG) to further improve metaphor detection \citep{reimann2025using, fuoli2026metaphor}. Related studies have also extended these approaches to multilingual, cross-lingual, and multimodal metaphor detection \citep{lai2023multilingual, hulsing2024cross, xu2024exploring}. Nevertheless, current results remain mixed, with the effectiveness of LLM-based approaches depending strongly on the task, prompting strategy, and evaluation dataset \citep{reimann2025using,fuoli2026metaphor,han2025dutch}.


Despite these advances, several challenges remain. Current systems often struggle with conventional metaphors, culturally specific metaphorical expressions, and implicit figurative meanings that require discourse-level reasoning or common-sense knowledge. Moreover, most existing work evaluates metaphor detection as an isolated classification task, while relatively little attention has been paid to how detected metaphorical expressions can support downstream applications such as machine translation and translation quality evaluation. 

Building upon our previous work on GPT-4-based metaphor detection in English news texts \cite{liang2025using}, 
this limitation motivates the \textit{first work package} of this PhD,
which extends metaphor identification towards \textit{multilingual} settings, improved prompting strategies, and more robust detection methods that can support downstream translation and evaluation.

\subsection{Metaphor Translation}

Metaphor translation has long been recognised as one of the most challenging problems in translation studies because metaphorical expressions often convey meaning beyond their literal lexical content. Unlike literal translation, successful metaphor translation requires preserving not only semantic equivalence but also conceptual mappings, communicative functions, stylistic effects, and cultural connotations. Classical translation scholars have therefore argued that metaphor translation should be viewed as a process of conceptual transfer rather than simple lexical substitution. \citet{newmark1988textbook}, for example, proposed a taxonomy of metaphor translation procedures ranging from reproducing the same metaphor in the target language to replacing or paraphrasing metaphorical expressions according to communicative needs. Similarly, \citet{vandenbroeck1981limits} and \citet{schaffner2004metaphor} emphasised that translation strategies should consider both linguistic form and underlying conceptual structures, highlighting the importance of cultural adaptation in cross-linguistic metaphor transfer.

The rapid development of neural machine translation (NMT) has substantially improved translation quality for general-purpose texts \cite{johnson-etal-2017-googles,kocmi2025findings}. Transformer-based systems and multilingual large language models (LLMs) are increasingly capable of generating fluent and contextually appropriate translations without explicit linguistic rules. Nevertheless, metaphorical language remains a persistent challenge because successful translation often requires pragmatic reasoning, cultural knowledge, and an understanding of implicit conceptual mappings \cite{han2026towards}. 
Existing MT systems may produce literal translations that distort metaphorical meaning, omit metaphorical imagery, or replace metaphors with more conventional expressions, thereby reducing their communicative impact \cite{liang2026metahope}.

Recent research has begun to investigate metaphor translation from a computational perspective. \citet{wang2024mmte} introduced MMTE, a multilingual benchmark specifically designed for evaluating machine translation of metaphorical language across several language pairs. Their study demonstrated that although modern LLM-based systems outperform earlier neural MT models, substantial performance gaps remain, particularly for culturally specific and conceptually complex metaphors. 
Other recent studies have similarly compared human translators with NMT systems and LLMs, showing that although modern LLMs often produce fluent and semantically adequate translations, they remain less reliable in preserving metaphorical expressions, conceptual mappings, and rhetorical effects, particularly for culturally specific or novel metaphors \citep{wang2024mmte,karakanta2025metaphors,li2025mind}. These findings suggest that improvements in general machine translation quality do not necessarily translate into reliable metaphor translation.

Although the quality of MT and LLM systems continues to improve rapidly, relatively little work has investigated how metaphor translation errors should be analysed systematically. Existing studies typically report overall translation quality or metaphor preservation rates, providing limited diagnostic information about the specific linguistic and conceptual problems underlying translation failures. Consequently, metaphor translation research increasingly requires evaluation frameworks capable of identifying fine-grained error types beyond conventional MT quality metrics. This observation motivates \textit{the second work package} of this PhD, which investigates metaphor-oriented translation evaluation and analysis.

\subsection{Translation Evaluation}

Automatic evaluation has become an indispensable component of modern machine translation research. Traditional corpus-level metrics such as BLEU \citep{papineni2002bleu}, METEOR \citep{banerjee-lavie-2005-meteor}, TER \cite{snover-etal-2006-study}, and LEPOR \cite{han-etal-2012-lepor,han2013language} primarily measure lexical or n-gram overlap between system outputs and reference translations. Although these metrics remain useful for benchmarking general translation quality, they are often insensitive to semantic equivalence and perform poorly when evaluating figurative language, where multiple valid translations may differ substantially in lexical realisation.

Recent evaluation methods have increasingly shifted towards semantic and learned metrics. COMET \citep{rei-etal-2020-comet} and BLEURT \citep{sellam-etal-2020-bleurt}, for example, employ pretrained language models to estimate translation quality based on semantic similarity rather than surface-form matching. These approaches generally correlate more strongly with human judgement than traditional automatic metrics and have become standard evaluation tools in contemporary MT research. Nevertheless, they remain general-purpose metrics and do not explicitly assess whether metaphorical meaning, conceptual mappings, or rhetorical effects have been preserved during translation.

To address the limitations of automatic metrics, human-centred evaluation frameworks have received growing attention. The Multidimensional Quality Metrics (MQM) framework \cite{lommel2024multi,lommel2014multidimensional,gladkoff2025non} provides a hierarchical taxonomy for classifying translation errors according to dimensions such as accuracy, fluency, terminology, and style. HOPE \cite{gladkoff2022hope} further simplified human evaluation by introducing a penalty-based annotation scheme that improves annotation efficiency and inter-annotator consistency while retaining diagnostic capability. However, neither MQM nor HOPE explicitly considers metaphor as an independent evaluation dimension.

Recent metaphor-specific evaluation research has therefore begun to bridge this gap. MMTE \cite{wang2024mmte} proposed benchmark datasets and dedicated evaluation metrics for metaphor translation, demonstrating that conventional automatic metrics frequently overestimate translation quality for figurative language. 
However, MMTE corpus has isolated sentences without context and it only contains verbal metaphors.
Building upon these developments, our previous 
work \cite{liang2026metahope} introduced MetaHOPE, a metaphor-oriented extension of the HOPE framework that incorporates \textit{five fine-grained error categories} to capture different aspects of metaphor translation quality, including communicative impact (IMP), 
adaptation to target language (RAM),
semantic accuracy (MIS), stylistic preservation (STL), and naturalness (PRF)·
MetaHOPE used context-aware translation before extracting the metaphor-containing segments for evaluation. It also produces a open-source parallel corpus during the post-editing phase that can be useful for future metaphor studies and benchmarks.
Preliminary pilot studies further showed that refined annotation guidelines substantially improve inter-annotator agreement, supporting the feasibility of reliable human evaluation for metaphor translation.

Despite these advances, metaphor evaluation remains largely dependent on manual annotation, making large-scale evaluation both expensive and time-consuming. Furthermore, current automatic metrics rarely incorporate information obtained during metaphor detection, and human evaluation frameworks generally assume that metaphorical expressions have already been identified. These limitations motivate the \textit{final work package} of this PhD, which investigates automatic metaphor translation quality estimation by integrating metaphor detection, human evaluation, and LLM-based evaluation into a unified computational framework.

\subsection{Research Questions from the Gap}

\textbf{Research Questions}
\begin{itemize}
    \item RQ1:
How can metaphorical expressions be reliably detected across languages?
\item RQ2:
How should metaphor translation quality be evaluated?
\item RQ3:
Can metaphor detection and translation evaluation be integrated into a unified computational framework?
\end{itemize}


\begin{figure*}[t]
\centering  \includegraphics[width=.9\textwidth]{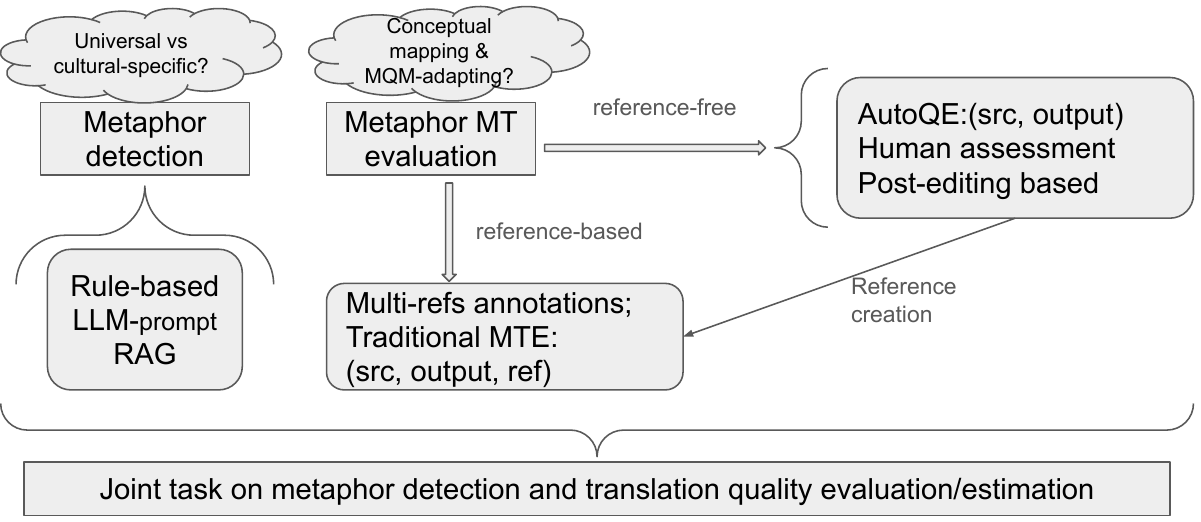}
  \caption{Thesis Proposal Framework on Metaphor Detection, Translation Evaluation, and Joint Modelling.}
  \label{fig:jj-proposal-fig}
\end{figure*}

\begin{figure*}[t]
\centering
\begin{tikzpicture}[
node distance=.3cm,
every node/.style={align=center,font=\small},
box/.style={
draw,
rounded corners,
minimum width=3.2cm,
minimum height=1cm,
fill=blue!8
},
arrow/.style={-Latex,thick}
]

\node[font=\Large\bfseries] (title) {Multilingual Metaphor Processing};

\node[box,below left=1cm and 1.5cm of title] (wp1)
{\textbf{WP1}\\Metaphor Detection};

\node[box,right=1.cm of wp1] (wp2)
{\textbf{WP2}\\Metaphor Translation};

\node[box,right=1.cm of wp2] (wp3)
{\textbf{WP3}\\Human Evaluation};

\node[box,right=1.cm of wp3] (wp4)
{\textbf{WP4}\\Automatic Evaluation};

\draw[arrow] (title) -- (wp1);
\draw[arrow] (title) -- (wp2);
\draw[arrow] (title) -- (wp3);
\draw[arrow] (title) -- (wp4);

\node[below=.2cm of wp1]
{Detection\\Framework};

\node[below=.2cm of wp2]
{MT \& LLM\\Translation\\Benchmark};

\node[below=.2cm of wp3]
{MetaHOPE\\Corpus\\Guidelines};

\node[below=.2cm of wp4]
{Automatic MQE\\LLM-as-a-Judge\\Joint Framework};

\end{tikzpicture}

\caption{Overview of the proposed PhD research programme. The four work packages form an end-to-end framework for multilingual metaphor processing, spanning metaphor detection, translation, human evaluation, and automatic evaluation.}
\label{fig:framework}
\end{figure*}
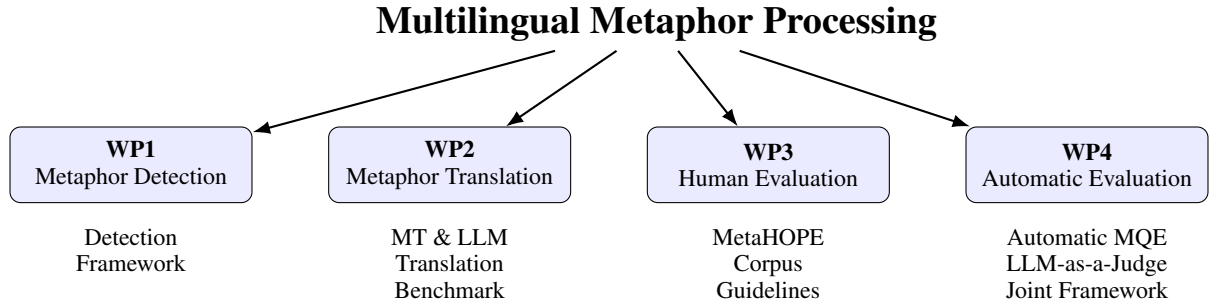

\section{Proposed Research Framework
}


We design the thesis framework in Figure \ref{fig:jj-proposal-fig}, which includes three components 1) metaphor detection, 2) metaphor translation evaluation, and 3) joint modelling on detection and translation evaluation.

Step I detection task: we plan to investigate rule-based, simple LLM-prompting with examples and chain-of-thoughts (CoTs), and RAG.

Step-II translation and evaluation task: we divide it into reference-free evaluation and reference-based. The reference-free evaluation includes 1) automatic quality estimation (QE) using source text and system output, 2) human native-speaker assessing the translation quality, and 3) post-editing based metrics.
The reference-based evaluation needs corpus with multiple-references for better mapping between hypothesis translation and the references. For this, we can use the created references from post-editing based evaluation from the reference-free step.

Step-III joint task: we investigate the possibility of joint modelling on both metaphor detection task and metaphor translation evaluation, which can be reference-based or reference-free quality estimation.

\section{WPs}
WP1 includes Detection,
datasets,
prompting,
fine-tuning, and 
multilingual models.


WP2 includes:
metaphor translation, 
MT/LLM, and
Translation Benchmark.

WP3 is on
Human Evaluation,
Integrating MetaHOPE \cite{liang2026metahope},
annotation,
corpus,
guidelines, and
reliability discussion.

WP4 is of
Automatic Eval,
auto MQE, 
llm-as-a-judge,
and end-to-end Joint framework (
Detection, Translation,
Evaluation
/Automatic QE,
LLM judge).

\section{Methodology}
\label{sec:method}

In our earlier work on LLM-based metaphor detection for English news texts \cite{liang2025using}, we discovered that GPT-4 achieved substantially lower performance than SOTA supervised metaphor detection models on conventional metaphor detection. We also observed that model performance was highly sensitive to prompt formulation. While one-shot prompting consistently improved performance over zero-shot prompting, providing additional examples did not lead to further improvements. 
Our error analysis further showed that prepositional metaphors (e.g., under in under strain) were particularly difficult to detect, with many prompts failing to identify such cases in the zero- to five-shot settings. 
Although detection improved in the ten-shot setting, our findings indicate that the observed gains cannot be straightforwardly interpreted as improved metaphor understanding and require further investigation. These findings motivate further research on prompt optimization, alternative output formats, fine-tuning open-source LLMs, and RAG.
Building upon our earlier work,
the first work package (\textbf{WP1}) extends metaphor identification towards \textit{multilingual} and more \textit{robust} detection methods. In particular, we will explore newer and more diverse open-source LLMs such as GPT -5, DeepSeek, Huanyuan, and Gemma 4 from 2025 onwards.
Furthermore, we will investigate chain of thoughts (CoTs) and RAG methods on this task, naming it \textit{MetaDetect}.

Regarding Metaphor Translation and Evaluation (WP2 and WP3), we are conducting the roll-out of larger size testing set annotation based on the MetaHOPE framework \cite{liang2026metahope} using 200+ segments with their context using 5 annotators for both translation directions on English-Chinese.
The resulting outputs from this will be a statistically robust analysis
of the current state-of-the-art LLMs regarding metaphor-related word translations. In addition, it will create a bilingual corpus to be shared publicly for research on English-Chinese pairs with multiple references (from multiple annotations).
We believe the multi-reference parallel corpus will also serve as an valuable resource
to the metaphor study field.

For WP4 on automatic evaluation, we are attending the WMT2026 shared task on Test Suites\footnote{\url{https://www2.statmt.org/wmt26/testsuite-subtask.html}}, where we used the AlphaMWE \cite{han2026towards,han-etal-2020-alphamwe} multilingual parallel corpus with MWE annotations including metaphors and idioms. We have received 30+ MT system submissions on the translation of this corpus that are being analysed using both automatic metrics and human evaluations. 
The automatic metrics being deployed include BLEU, CharF++, LEPOR, and BERTscore, covering both lexical-based and word-embedding metrics.

Our next step on the thesis moving forward is about end-to-end metaphor detection and translation evaluation modelling, i.e., part of WP4. 
For the end-to-end models, we will investigate ``LLMs as Agents'' \cite{plaat2025agentic}, applying different LLMs for detection and translation evaluation from two joint phases in a collective manner. 
LLM-as-a-judge will also be explored as the evaluation phase with human-in-the-loop, as suggested from the literature \cite{wit2026measuringpracticeshareddecisionmaking}.

Overall, the four WPs are coherently connected: 
outputs of WP1 feed into WP2;
WP2 provides resources for WP3; and
WP3 supervises WP4.

\section{Research Plans}
\begin{figure*}[th!]
\centering

\begin{tikzpicture}[
    >=Latex,
    font=\small,
    milestone/.style={
        draw,
        rounded corners,
        fill=blue!8,
        minimum width=2.5cm,
        minimum height=0.9cm,
        align=center
    }
]

\draw[thick,->] (0,0) -- (12.5,0);

\node[milestone] (m1) at (1.5,0.8)
{\textbf{Milestone 1}\\Metaphor Detection};

\node[milestone] (m2) at (4.5,0.8)
{\textbf{Milestone 2}\\Translation};

\node[milestone] (m3) at (7.8,0.8)
{\textbf{Milestone 3}\\Automatic Evaluation};

\node[milestone] (m4) at (11.2,0.8)
{\textbf{Milestone 4}\\Joint Framework\\+ Thesis};

\draw (1.5,0) -- (1.5,0.45);
\draw (4.5,0) -- (4.5,0.45);
\draw (7.8,0) -- (7.8,0.45);
\draw (11.2,0) -- (11.2,0.45);

\node[below] at (1.5,-0.1) {Phase 1};
\node[below] at (4.5,-0.1) {Phase 2};
\node[below] at (7.8,-0.1) {Phase 3};
\node[below] at (11.2,-0.1) {Phase 4};

\end{tikzpicture}

\caption{Proposed research timeline and milestones for the PhD project.}
\label{fig:timeline}

\end{figure*}
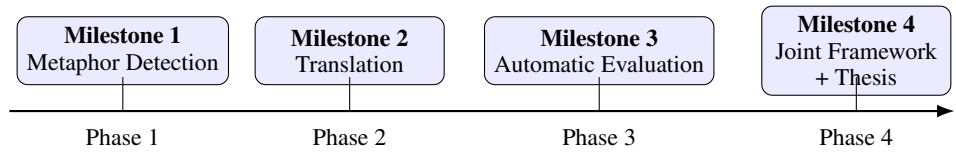

The Thesis Research Plan is listed in Figure \ref{fig:timeline}.

\section{Expected Contributions}
Expected outputs include:

\begin{itemize}
\item Resources
\begin{itemize}
    \item multilingual corpus
    \item benchmark
    \item annotation guidelines
\end{itemize}
\item Methods
\begin{itemize}
    \item MetaHOPE
    \item automatic evaluator
    \item joint framework
\end{itemize}
\item Scientific contributions
\begin{itemize}
    \item end-to-end metaphor processing paradigm
\end{itemize}
\end{itemize}

\section{Conclusion}

This PhD proposal presents a research agenda towards end-to-end multilingual metaphor processing by integrating metaphor detection, metaphor translation, human evaluation, and automatic translation evaluation into a unified computational framework. While these research areas have each advanced considerably in recent years, they have largely been investigated independently. As a result, current multilingual NLP systems still face substantial challenges when processing metaphorical language, particularly across languages and cultures.

The proposed research addresses this gap through four complementary work packages. The first investigates robust multilingual metaphor detection using both linguistic knowledge and recent large language models. The second examines metaphor translation and develops benchmark resources for analysing metaphor translation strategies across machine translation and LLM systems. The third develops a reliable human evaluation methodology for metaphor translation, including annotation guidelines, corpora, and empirical studies of annotation reliability. Finally, the fourth explores automatic metaphor translation evaluation by integrating metaphor detection, human evaluation, and LLM-based quality estimation into a unified framework.

Beyond developing individual methods and datasets, this research aims to establish stronger connections between metaphor detection and translation evaluation. Information produced during metaphor detection has the potential to support more accurate translation quality estimation, while human evaluation can provide supervision for developing metaphor-aware automatic evaluation models. By investigating these interactions, the proposed research seeks to move beyond isolated task-specific solutions towards a coherent framework for multilingual metaphor processing.

The expected outcomes include new multilingual resources, reproducible annotation methodologies, benchmark datasets, and automatic evaluation techniques that facilitate future research on figurative language processing. More broadly, the proposed work contributes to the development of multilingual NLP systems that better capture the linguistic, conceptual, and cultural dimensions of metaphor, thereby supporting more reliable machine translation and evaluation of figurative language.

\section*{Current Progress}

\begin{itemize}
    \item Completed: GPT-4 metaphor detection study; MetaHOPE framework.
    \item In progress: Large-scale MetaHOPE annotation; WMT 2026 Test Suites evaluation.
    \item Planned: Joint metaphor detection and evaluation models; LLM-as-a-judge framework.
\end{itemize}




\bibliography{custom}

\appendix

\section{Initial Annotation Guidelines}
\label{sec:appendix}

Two files: 
``Annotation Guideline for The Categorization of Potentially Universal Metaphors and Culture''
and 
``Annotation Guidelines for Metaphor Translation Strategies and Quality Evaluation''
both available at \url{https://github.com/Jiahui84/MetaHOPE}

\end{document}